\documentclass[10pt,twocolumn,letterpaper]{article}

\usepackage[pagenumbers]{cvpr} 
\usepackage{times}
\usepackage{subfloat}
\usepackage{multirow}

\usepackage{graphicx}
\usepackage{amsmath}
\usepackage{amssymb}
\usepackage{booktabs}

\usepackage[pagebackref,breaklinks,colorlinks]{hyperref}

\usepackage[capitalize]{cleveref}
\crefname{section}{Sec.}{Secs.}
\Crefname{section}{Section}{Sections}
\Crefname{table}{Table}{Tables}
\crefname{table}{Tab.}{Tabs.}

\begin{document}

\title{X-Pruner: eXplainable Pruning for Vision Transformers}

\author{Lu Yu\\
James Cook University\\
	{\tt\small lu.yu@my.jcu.edu.au}
	\and
	Wei Xiang\thanks{Corresponding author.}\\
	La Trobe University\\
	{\tt\small w.xiang@latrobe.edu.au}
}

\maketitle

\begin{abstract}
  Recently vision transformer models have become prominent models for a range of tasks. These models, however, usually suffer from intensive computational costs and heavy memory requirements, making them impractical for deployment on edge platforms. Recent studies have proposed to prune transformers in an unexplainable manner, which overlook the relationship between internal units of the model and the target class, thereby leading to inferior performance. To alleviate this problem, we propose a novel explainable pruning framework dubbed X-Pruner, which is designed by considering the explainability of the pruning criterion. Specifically, to measure each prunable unit's contribution to predicting each target class, a novel explainability-aware mask is proposed and learned in an end-to-end manner. Then, to preserve the most informative units and learn the layer-wise pruning rate, we adaptively search the layer-wise threshold that differentiates between unpruned and pruned units based on their explainability-aware mask values. To verify and evaluate our method, we apply the X-Pruner on representative transformer models including the DeiT and Swin Transformer. Comprehensive simulation results demonstrate that the proposed X-Pruner outperforms the state-of-the-art black-box methods with significantly reduced computational costs and slight performance degradation. Code is available at \url{https://github.com/vickyyu90/XPruner}.
\end{abstract}

\section{Introduction}
\label{sec:intro}
Over the last few years, transformers have attracted increasing attention in various challenging domains, such as natural language processing, vision, or graphs \cite{Chavan2022, dosovitskiy2020ViT}. It is composed of two key modules, namely the Multi-Head Attention (MHA) and Multi-Layer Perceptron (MLP). However, similar to CNNs, the major limitations of transformers include the gigantic model sizes with intensive computational costs. Which severely restricts their deployment in resource-constrained devices like edge platforms. To compress and accelerate transformer models, a variety of techniques naturally emerge. Popular approaches include weight quantization \cite{ZhangZLBHCZ22}, knowledge distillation \cite{0004HWCCC22}, filter compression \cite{TangWXTX0020}, and model pruning \cite{PanZLH021}. Among them, model pruning especially structured pruning has gained considerable interest that removes the least important parameters in pre-trained models in a hardware-friendly manner, which is thus the focus of our paper. 

Due to the significant structural differences between CNNs and transformers, although there is prevailing success in CNN pruning methods, the research on pruning transformers is still in the early stage. Existing studies could empirically be classified into three categories. (1) Criterion-based pruning resorts to preserving the most important weights/attentions by employing pre-defined criteria, e.g., the L1/L2 norm \cite{PanPJWFO21}, or activation values \cite{ChenCGYZW21}. (2) Training-based pruning retrains models with hand-crafted sparse regularizations \cite{ZhuZW21} or resource constraints \cite{0004HWCCC22, YuCSYTY0W22}. (3) Architecture-search pruning methods directly search for an optimal sub-architecture based on pre-defined policies \cite{FanGJ20, Chavan2022}. Although these studies have made considerable progress, two fundamental issues have not been fully addressed, \ie, the optimal layer-wise pruning ratio and the weight importance measurement.

For the first issue, the final performance is notably affected by the selection of pruning rates for different layers. To this end, some relevant works have proposed a series of methods for determining the optimal per-layer rate \cite{ChenFC0ZWC20, FrantarA22}. For instance, Michel \etal \cite{MichelLN19} investigate the effectiveness of attention heads in transformers for NLP tasks and propose to prune attention heads with a greedy algorithm. Yu \etal \cite{0004HWCCC22} develop a pruning algorithm that removes attention scores below a learned per-layer threshold while preserving the overall structure of the attention mechanism. However, the proposed methods do not take into account the inter-dependencies between weight. Recently, Zhu \etal \cite{ZhuZW21} introduce the method VTP with a sparsity regularization to identify and remove unimportant patches and heads from the vision transformers. However, VTP needs to try the thresholds manually for all layers.

For the second issue, previous studies resort to identifying unimportant weights by various importance metrics, including magnitude-based, gradient-based \cite{HuangWCW19, NaMCY21}, and mask-based \cite{WangWL20}. Among them, the magnitude-based approaches usually lead to suboptimal results as it does not take into account the potential correlation between weights \cite{TangWXTX0020}. In addition, gradient-based methods often tend to prune weights with small values, as they have small gradients and may not be identified as important by the backward propagation. Finally, the limitation of current mask-based pruning lies in two folds: (1) Most mask-based pruning techniques manually assign a binary mask w.r.t. a unit according to a per-layer pruning ratio, which is inefficient and sub-optimal. (2) Most works use a non-differentiable mask, which results in an unstable training process and poor convergence.

In this paper, we propose a novel explainable structured pruning framework for vision transformer models, termed X-Pruner, by considering the explainability of the pruning criterion to solve the above two problems. As stated in the eXplainable AI (XAI) field \cite{arrieta2020explainable}, important weights in a model typically capture semantic class-specific information. Inspired by this theory, we propose to effectively quantitate the importance of each weight in a class-wise manner. Firstly, we design an explainability-aware mask for each prunable unit (e.g., an attention head or matrix in linear layers), which measures the unit's contribution to predicting every class and is fully differentiable. Secondly, we use each input's ground-truth label as prior knowledge to guide the mask learning, thus the class-level information w.r.t. each input will be fully utilized. Our intuition is that if one unit generates feature representations that make a positive contribution to a target class, its mask value w.r.t. this class would be positively activated, and deactivated otherwise. Thirdly, we propose a differentiable pruning operation along with a threshold regularizer. This enables the search of thresholds through gradient-based optimization, and is superior to most previous studies that prune units with hand-crafted criteria. Meanwhile, the proposed pruning process can be done automatically, \ie, discriminative units that are above the learned threshold are retained. In this way, we implement our layer-wise pruning algorithm in an explainable manner automatically and efficiently. In summary, the major contributions of this paper are:
\begin{itemize}
  \item We propose a novel explainable structured pruning framework dubbed X-Pruner, which prunes units that make less contributions to identifying all the classes in terms of explainability. To the best knowledge of the authors, this is the first work to develop an explainable pruning framework for vision transformers;
  \item We propose to assign each prunable unit an explainability-aware mask, with the goal of quantifying its contribution to predicting each class. Specifically, the proposed mask is fully differentiable and can be learned in an end-to-end manner;
  \item Based on the obtained explainability-aware masks, we propose to learn the layer-wise pruning thresholds that differentiate the important and less-important units via a differentiable pruning operation. Therefore, this process is done in an explainable manner;
  \item Comprehensive simulation results are presented to demonstrate that the proposed X-Pruner outperforms a number of state-of-the-art approaches, and shows its superiority in gaining the explainability for the pruned model.
\end{itemize}

\begin{figure*}[t]
  \setlength{\belowcaptionskip}{0pt}
  \setlength{\abovecaptionskip}{5pt}
  \begin{center}
  \includegraphics[width=0.7\linewidth]{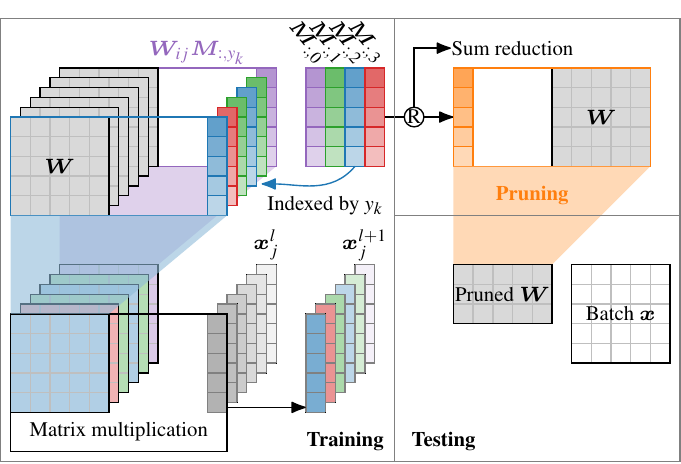}
\caption{Pipeline of our proposed X-Pruner framework. We first train a transformer with the proposed explainability-aware masks, with the goal of quantifying each unit's contribution to predicting each class. Then we explore the layer-wise pruning threshold under a pre-defined cost constraint. Finally, a fine-tune procedure is executed for the pruned model.}
  \end{center}
\label{fig1}
\end{figure*}

\section{Related Work}
\label{sec2}
\subsection{Pruning for transformers}
Pruning has been a popular approach for removing the least important weights in transformer models. The existing methods can be mainly categorized into unstructured and structured pruning. For unstructured pruning, techniques such as magnitude-based and hessian-based have been proposed \cite{ChenFC0ZWC20, Sanh0R20}. However, they result in irregular sparsity, causing sparse tensor computations that are difficult to align with hardware efficiency.

The above problem can be alleviated by structured pruning, where uninformative contiguous structures of a pre-trained model such as attention heads, rows of weight matrix, are removed. For instance, Michel \etal \cite{MichelLN19} found that a large percentage of attention heads can be pruned without scarifying much performance. Fan \etal \cite{FanGJ20} proposed a structured dropout, which selects sub-structures of a model during the inference time. Wang \etal \cite{WangWL20} pruned rank-1 components inside large language models using a parameterization method. Liu \etal \cite{LiuLY21} assembled several model compression techniques on a range of pre-trained language models, and gained impressive results.However, these works focus on pruning transformers for NLP tasks. 

For vision transformers, Chen \etal \cite{ChenCGYZW21} explored unstructured and structured sparsity, and proposed a first-order importance approximation method to remove attention heads. Recently, Yu \etal \cite{0004HWCCC22} propose a structured pruning method for vision transformers, which involves a 0/1 mask that differentiates unimportant/important parameters based on the magnitude of the model parameters. Although it uses a differentiable threshold, the mask is non-differentiable, which could cause the gradients to be biased and result in suboptimal results of the remaining weights. Yu \etal \cite{YuCSYTY0W22} proposed to integrate three efficient approaches including pruning, layer skipping, and knowledge distillation into a unified framework to produce a compact transformer. Although these existing methods have made significant advances, the designing of importance metrics remains an open problem to explore.

\subsection{XAI techniques for transformers}
In terms of XAI approaches, there are a variety of recent studies aiming to explain predictions made by transformers. Chefer \etal \cite{CheferLRP21} proposed a layer-wise relevance propagation (LRP) method that applies to both positive and negative contributions. This approach, however, is not able to provide the interpretation for attention modules besides self-attention. Voita \etal \cite{voita2019analyzing} found that only a small portion of heads have a salient impact on model decisions. Jawahar \etal \cite{JawaharSS19} proved that semantic patterns are gained through higher layers. Abnar \etal \cite{AbnarZ20} proposed to combine the attention scores across multiple layers, but this method failed to distinguish between positive and negative attributions. Raghu \etal \cite{RaghuUKZD21} analyzed the internal representations of vision transformers and found out that they learn more uniform representations across all layers. Recently, Chefer \etal \cite{CheferGW21} also proposed a generic approach to explain transformers including multi-modal ones. The advances in XAI technology have brought about substantial improvements to wide-ranging tasks. The capability of explainability in guiding model pruning, however, remains unexplored in the literature. Therefore, we propose the X-Pruner that aims to make full use of model explainability to derive the importance metric for model pruning.

\section{Methodology}
\label{sec3}
\subsection{Problem definition}
Our proposed X-Pruner aims to explore structured pruning by removing prunable units (e.g., rows of weight matrix and attention heads) in vision transformers. Let $\mathcal{\boldsymbol{D}}$ be a training dataset, which consists of $N$ training pairs $\{(\boldsymbol{x}_1,\boldsymbol{y}_1),...,(\boldsymbol{x}_N,\boldsymbol{y}_N)\}$. Considering an $L$-layer transformer $f(\boldsymbol{W})$, its parameters are represented by $\boldsymbol{W}=(\boldsymbol{W}^1,\boldsymbol{W}^2,...,\boldsymbol{W}^L)$, where $\boldsymbol{W}^l \in \mathbb{R}^{d_l}, 1 \leq l \leq L$, $d_l$ is the number of prunable parameters in the $l$-th layer. Given a target pruning ratio $\alpha$, the pruning process can be regarded as the form of layer-wise operation with pruning rates $\{r_l\}_{l=1}^L$:
\begin{align}
    (r_1,r_2,...,r_L)^* & = {\rm argmin}\mathcal{L}(f(\boldsymbol{W};r_1,r_2,...,r_L;\boldsymbol{x}), \boldsymbol{y}), \nonumber \\
    s.t. & \quad \frac{P(f(\boldsymbol{W};\{r_l\}))}{P(f(\boldsymbol{W}))} \geq \alpha,
    \label{eq1}
\end{align}
where $r_l$ is the $l$-th layer's pruning rate, and $P(\cdot)$ is a resource evaluation metric.

\subsection{The Proposed X-Pruner}
\subsubsection{Explainability-aware mask}
To fully utilize the class-level information, we propose to assign each prunable unit an explainability-aware mask, which is used to quantify the contribution of each unit to identifying every class. Specifically, the proposed mask is a class-level mask for each class instead of a scalar mask for all classes. For instance, given the weights in the $l$-th self-attention layer consists of query $\boldsymbol{W}_l^Q \in \mathbb{R}^{n \times d}$, key $\boldsymbol{W}_l^K \in \mathbb{R}^{n \times d}$, and value $\boldsymbol{W}_l^V \in \mathbb{R}^{n \times d}$, where $n$ and $d$ are the number of input and output dimension. The mask for head $h$ is formulated as $\boldsymbol{M}_{l,h}^H \in \mathbb{R}^{C \times d}$, where $C$ is the total number of classes. That is to say, $\boldsymbol{M}_{l,h,i}^H$ is built to quantify the contribution of head $h$ for recognizing the $i$-th class. Evidently, a scalar mask used in prior works is a special case of our method where values of $\boldsymbol{M}_{l,h,i}^H$ are the same. Thus, given input $\boldsymbol{x}_i$ with its class label $\boldsymbol{y}_i$, to apply the mask, the product between weight and its corresponding mask is performed. That is, the self-attention operation for head $h$ can be expressed as follows:
\begin{align}
    & \boldsymbol{\alpha}_{l,h} = S(\frac{(\boldsymbol{W}_{l,h}^Q \boldsymbol{x}_i)^T \boldsymbol{W}_{l,h}^K \boldsymbol{x}_i}{\sqrt{d}}), \\
    & {\rm Attn}_{l,h}(\boldsymbol{x}) = \boldsymbol{\alpha}_{l,h} \boldsymbol{W}_{l,h}^V \boldsymbol{x}_i, \\
    & {\rm MHA}(\boldsymbol{x}, \boldsymbol{M}_{l}^H) = {\sum}_{h=1}^{H} \boldsymbol{M}_{l, h, \boldsymbol{y}_i}^H {\rm Attn}_{l,h}(\boldsymbol{x}),
\end{align}
where $S(\cdot)$ is the softmax function, $\boldsymbol{\alpha}_h$ is the $h$-th attention weight, and $H$ is the total number of attention heads. 

Meanwhile, we apply the similar idea to the MLP and other linear projection layers. Let us denote the weight matrix in a linear layer by $\boldsymbol{W}_l \in \mathbb{R}^{m \times n}$, where the $m$ and $n$ are the dimensions. Its corresponding mask is defined by $\boldsymbol{M}_l^F \in \mathbb{R}^{C \times m \times n}$. Then, the feed forward process in the linear layer is expressed as:
\begin{align}
    {\rm FC}(\boldsymbol{Z}_l, \boldsymbol{M}_{l}^F) = \boldsymbol{M}_{l,\boldsymbol{y}_i} \boldsymbol{W}_l^F \boldsymbol{Z}_l,
\end{align}
where $\boldsymbol{Z}_l$ is the input to the $l$-th layer. We omit the bias across all layers for simplicity.

Recall that our explainability-aware mask aims to identify weights influential to the predicted label. As such, it is desirable for mask $\boldsymbol{M}_{:,c}$ to vary slowly if input images all belong to the same class $c$, rendering a smooth explainability-aware mask. Therefore, we propose to add a smoothness-aware constraint for the mask. More specifically, we take the second derivative of the mask values w.r.t. the input and predicted class, and choose its $L_1$ norm as the smoothness-aware constraint:
\begin{align}
    \mathcal{L}_{\rm smooth}(\boldsymbol{M}) = \sum_{l=1}^L \sum_{c=1}^{C}|\nabla^2 \boldsymbol{M}_{:,c}^l|_1.
\end{align}

Moreover, to address the issue of redundancy among the prunable units, rather than declaring all units as relevant to the model's prediction, we impose the following sparsity constraint on the masks:
\begin{align}
    \mathcal{L}_{\rm sparse}(\boldsymbol{M}) = \sum_{l=1}^L \sum_{c=1}^{C}||\boldsymbol{M}_{:,c}^l||_2.
\end{align}

Overall, the total loss function is defined as follows:
\begin{align}
    \mathcal{L}_{\rm total} = \mathcal{L}_{\rm ce} + \lambda_{\rm sm} \mathcal{L}_{\rm smooth} + \lambda_{\rm sp} \mathcal{L}_{\rm sparse}(\boldsymbol{M}),
\end{align}
where $\mathcal{L}_{\rm ce}$ is the cross-entropy loss, $\lambda_{\rm sm}$ and $\lambda_{\rm sp}$ are the hyperparameters.

Unlike prior works that use a binary mask to quantify the contribution of each unit for all classes, we propose to capture the importance of every unit w.r.t. each class with a differentiable mask. After training, the sum value of each learned mask explicitly denotes its contribution to identifying all classes. In this way, our learned explainability-aware masks gain the representation ability for revealing the inner reasoning process in transformers in an end-to-end manner, which essentially offers a global examination of the importance of every single unit in an intuitively explainable manner. Noticeably, the weights of the pre-trained model remain fixed during the training procedure. Therefore, we empirically observe that only a few epochs are required to train our proposed explainability-aware mask.

\subsubsection{Explainable pruning}
The goal of the proposed X-Pruner is to preserve the most important units for identifying target classes in a pruned model. This is achieved by removing units with the least-impact explainability-aware masks. Previous works resort to measuring the importance of individual units with a manually chosen per-layer threshold, which is computationally intractable as the parameter search space is exhaustive \cite{YuCSYTY0W22}. In this work, we propose to learn the layer-wise threshold by designing a differentiable pruning operation along with a threshold regularizer, which is superior to most prior studies with better control over the non-uniform sparsity.

Intuitively, with the obtained explainability-aware masks, the less-important units with mask values below a certain threshold should be pruned, while important ones are preserved. However, most of current approaches use a manually selected threshold, which is difficult to optimize in a trainable process. To tackle this issue, we propose a differentiable pruning operation for explainability-aware masks. Mathematically, the differentiable pruning operation is expressed as follows:
\begin{equation}
    \hat{\boldsymbol{M}}^l = \left\{
        \begin{aligned}
            & \boldsymbol{M}^l \tanh(n(\boldsymbol{M}^l-\theta^l)), \enspace \boldsymbol{M}^l \in \Phi(\boldsymbol{M}^l|1-r^l), \\
            & p \tanh(n(\boldsymbol{M}^l-\theta^l)), \enspace \quad {\rm otherwise},
        \end{aligned}
        \right. 
\end{equation}
where $r^l$ is the pruning ratio for layer $l$, and $\Phi(\boldsymbol{M}^l|1-r^l)$ is a function that returns the top $(1-r^l)\%$ sorted elements in $\boldsymbol{M}^l$. With a proper setting of $n$ and $p$, the value of $\tanh(\cdot)$ asymptotically approaches one for $\boldsymbol{M}^{l} \in \Phi(\boldsymbol{M}^l|1-r^l)$, which results in $\hat{\boldsymbol{M}}^l \approx \boldsymbol{M}^{l}$. In that case, our proposed differentiable pruning operation implies that discriminative units that contribute more to identifying classes above an adaptive threshold are retained, while those that contribute less are suppressed. By assigning a large positive value to $n$, our proposed pruning function enables learning the threshold $\theta^l$ with the backward gradient. In our experiments, we empirically verify that letting $p = 500$ and $n = 10$ guarantees a stable training process and yields good results for pruning.

Subsequently, we compute the accumulated pruning rate $R$ across all prunable layers as follows:
\begin{equation}
    R = \sum_{l=1}^L \frac{r^l*n^l}{N},
\end{equation}
where $n^l$ represents the total prunable parameters of the layer $l$ and $N$ denotes the number of all unpruned parameters. 

To learn the layer-wise pruning rate with the given pruning rate $\alpha$ in an end-to-end manner, we propose a novel regularization term $\mathcal{L}_{\rm R}$ in the augmented Lagrangian method, which converts the optimization problem in \cref{eq1} to an unconstrained penalized expression. Specifically, it is expressed as
\begin{equation}
    \mathcal{L}_{\rm R} = \beta (\alpha - R)^2 + \gamma (\alpha - R),
        \label{eq11}
\end{equation}
where $\beta$ and $\gamma$ are trainable parameters, and the unconstrained problem of \cref{eq11} can be solved using gradient descent-based techniques. Overall, the total loss function for the proposed X-Pruner is given by
\begin{equation}
    \mathcal{L} = \mathcal{L}_{\rm ce} + \mathcal{L}_{\rm R}.
    \label{eq12}
\end{equation}
The optimization problem of \cref{eq12} allows us to lift up units with discriminative masks that are important to the model decisions while suppressing less-important ones. Moreover, it implies that the layer-wise pruning rate $r^l$ tends to be larger when it has larger $n^l$, which is natural for exploiting the dynamic sparsity across all layers.

After training, we accordingly remove the least-impact units with the learned pruning rate $\{r^1,r^2,...,r^L\}$, and integrate the left explainability-aware masks $\boldsymbol{M}$ into the pruned model per layer by setting $\boldsymbol{W} = \boldsymbol{W} \ast \boldsymbol{M}$, and further fine-tune the pruned model. In summary, our proposed explainable pruning method X-Pruner that is capable of identifying and preserving important units in an explainable and trainable way, which overcomes the drawbacks of existing black-box pruning methods and provides empirical guarantees on the accuracy of the pruned model.

\section{Experiments}
\label{sec:experi}
To evaluate the performance of the X-Pruner, we conduct experiments on the CIFAR-10 \cite{Krizhevsky2009} and ILSVRC-12 datasets \cite{DengDSLL009}. CIFAR-10 includes 10 classes, consisting of 50K training and 10K validation images. ILSVRC-12 contains images of 1K classes, and its training and validation sets have 1.28M images and 50K images, respectively. For a fair comparison with existing methods, we prune the DeiT \cite{TouvronCDMSJ21} and Swin Transformer \cite{liu2021swin} architectures on classification tasks \cite{YuCSYTY0W22,0004HWCCC22}. Additionally, we conduct a series of ablation studies to discover the performance contribution from different components in our framework.

\subsection{Implementation details}
All experiments are implemented using PyTorch on NVIDIA Tesla V100 GPUs. We use pre-trained weights to initialize vision transformer models and use them as baseline models. During the training process for explainability-aware masks, the learning rate is set to be 0.01 with a batch size of 128, and we use the SGD optimizer with momentum 0.9. Empirically, the mask training process is 50 epochs for the DeiT and 30 epochs for the Swin Transformer. Which takes around 300 V100 GPU hours. In the explainable pruning process, we initially set all $r^l$ to $\alpha$. The learning rate for $\theta^l$ and $r^l$ is set to be 0.02 and fine-tuned with the AdamW optimizer. The learning rate for the other parameters and momentum are 5 $\times {10}^{-4}$ and 0.9, respectively. The DeiT models are trained for 80 epochs and Swin Transformers are trained for 30 epochs. We follow the training strategies used in the original DeiT and Swin Transformers \cite{liu2021swin} except knowledge distillation. $\beta$ and $\gamma$ are initialized to be zero and then optimized during training.

\begin{table*}
  \setlength{\belowcaptionskip}{0pt}
  \setlength{\abovecaptionskip}{-5pt}
    \caption{Comparison with the state-of-the-art methods on the ILSVRC-12 dataset. FLOPs remained denotes the remained ratio of FLOPs to the full-model FLOPs. $*$ indicates utilizing knowledge distillation in the training process.}
    \begin{center}
    \setlength{\tabcolsep}{2pt}
    \begin{tabular}{l|l|c|c|c|c}
    \toprule  
    Model & Method & Top-1 Acc. (\%) & Top-5 Acc. (\%) & FLOPs (G) & FLOPs remained (\%) \\
    \midrule  
    \multirow{6}{*}{DeiT-T} & Baseline & 72.2 & 91.10 & 1.3 & 100 \\
    & (NeurIPS'20) SCOP \cite{TangWXTX0020} & 68.9 & 89.00 & 0.8 & 61.5 \\
    & (ICCV'21) HVT \cite{PanZLH021} & 69.7 & 89.40 & 0.7 & 53.8 \\
    & (ICLR'22) UVC$^*$ \cite{YuCSYTY0W22} & 70.6 & - & 0.5 & 39.1 \\
    & (AAAI'22) WDPruning \cite{0004HWCCC22} & 70.3 & 89.82 & 0.7 & 53.8 \\
    & \bf{X-Pruner} & \bf{71.1} & \bf{90.11} & \bf{0.6} & \bf{49.2} \\
    \midrule
    \multirow{6}{*}{DeiT-S} & Baseline & 79.8 & 95.00 & 4.6 & 100 \\
    & (NeurIPS'20) SCOP \cite{TangWXTX0020} & 77.5 & 93.50 & 2.6 & 56.5 \\
    & (ICCV'21) HVT \cite{PanZLH021} & 78.0 & 93.83 & 2.4 & 52.2 \\
    & (ICLR'22) UVC$^*$ \cite{YuCSYTY0W22} & 78.82 & - & 2.3 & 50.4 \\
    & (AAAI'22) WDPruning \cite{0004HWCCC22} & 78.38 & 94.05 & 2.6 & 56.5 \\
    & \bf{X-Pruner} & \bf{78.93} & \bf{94.24} & \bf{2.4} & \bf{52.1} \\
    \midrule
    \multirow{5}{*}{DeiT-B} & Baseline & 81.8 & 95.59 & 17.6 & 100 \\
    & (NeurIPS'20) SCOP \cite{TangWXTX0020} & 79.7 & 94.50 & 10.2 & 58.3 \\
    & (ICLR'22) UVC$^*$ \cite{YuCSYTY0W22} & 80.57 & - & 8.0 & 45.5 \\
    & (AAAI'22) WDPruning \cite{0004HWCCC22} & 80.76 & 95.36 & 9.9 & 56.3 \\
    & \bf{X-Pruner} & \bf{81.02} & \bf{95.38} & \bf{8.5} & \bf{48.5} \\
    \bottomrule
    \end{tabular}
  \end{center}
\label{tab1}
\end{table*}

\begin{figure*}
  \setlength{\abovecaptionskip}{-0.5cm}
  \setlength{\belowcaptionskip}{-0.2cm}
  \begin{center}
    \subfloat{
        \begin{minipage}[h]{0.03\textwidth}
     \rotatebox{90}{~~~~~~~~~~\scalebox{0.9}[0.9]{Input Image}}
   \rotatebox{90}{~~~~~~~\scalebox{0.9}[0.9]{Full Model}}
    \rotatebox{90}{~~~~~~\scalebox{0.9}[0.9]{WDPruning \cite{0004HWCCC22}}}
    \rotatebox{90}{~~~~~~~~\scalebox{0.9}[0.9]{UVC \cite{YuCSYTY0W22}}~~~~}
    \rotatebox{90}{~~~~~~~~~~~~\scalebox{0.9}[0.9]{X-Pruner (Ours)}}
        \end{minipage}}
        \hspace{-6pt}
  \subfloat{
        \begin{minipage}[h]{0.19\textwidth}
     \includegraphics[width=0.95\textwidth,height=0.75\textwidth]{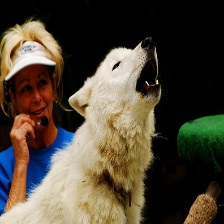}
    \includegraphics[width=0.95\linewidth, height=0.75\linewidth]{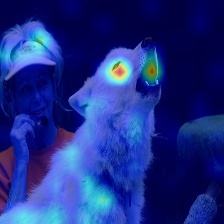}
   \includegraphics[width=0.95\linewidth, height=0.75\linewidth]{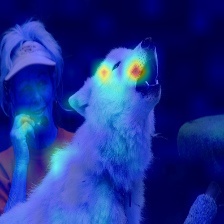}
    \includegraphics[width=0.95\linewidth, height=0.75\linewidth]{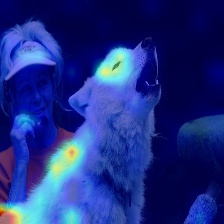}
    \includegraphics[width=0.95\linewidth, height=0.75\linewidth]{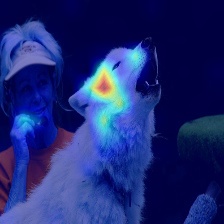}
    \centerline{White wolf}
    \label{fig:short-4a}
        \end{minipage}}
        \hspace{-5pt}
  \subfloat{
          \begin{minipage}[h]{0.19\textwidth}
       \includegraphics[width=0.95\textwidth,height=0.75\textwidth]{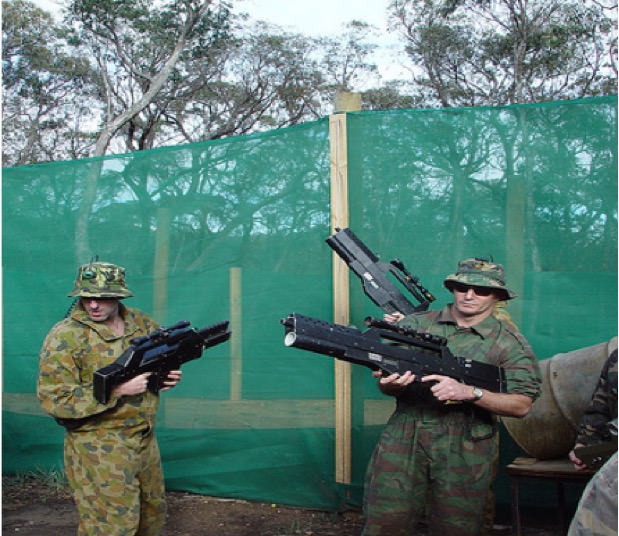}
      \includegraphics[width=0.95\linewidth, height=0.75\linewidth]{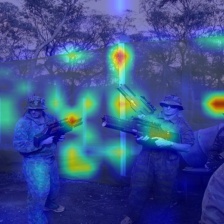}
     \includegraphics[width=0.95\linewidth, height=0.75\linewidth]{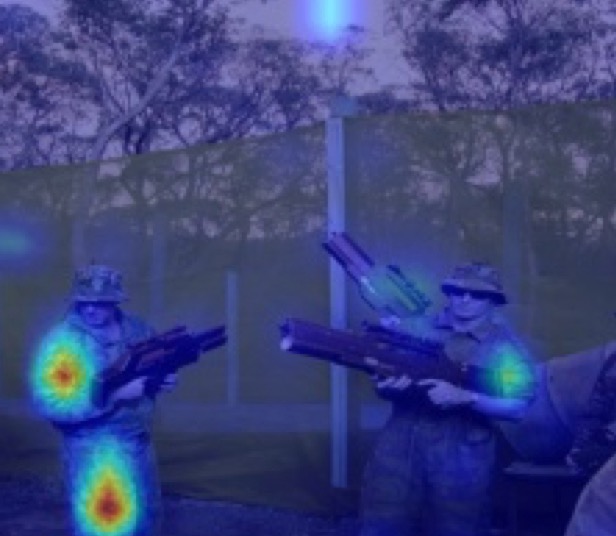}
      \includegraphics[width=0.95\linewidth, height=0.75\linewidth]{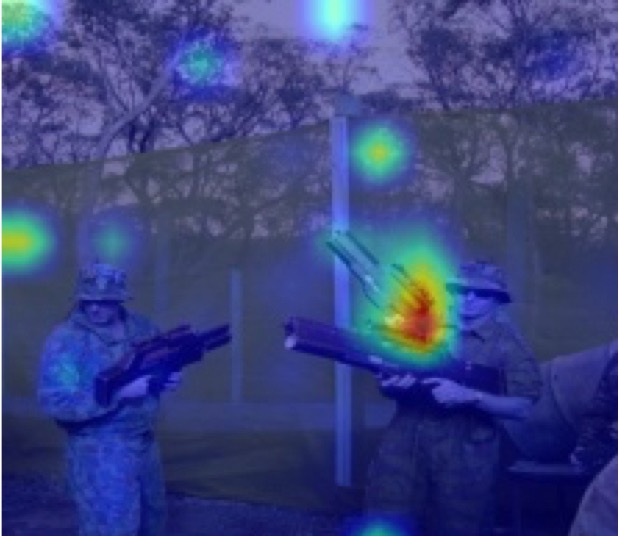}
      \includegraphics[width=0.95\linewidth, height=0.75\linewidth]{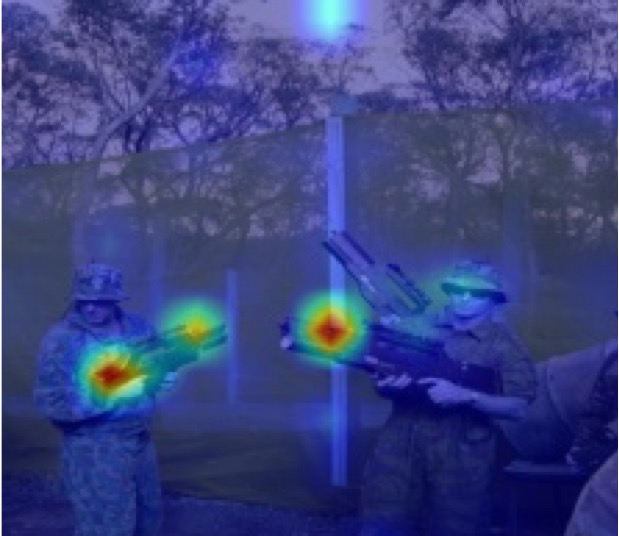}
      \centerline{Rifle}
      \label{fig:short-4c}
          \end{minipage}}
          \hspace{-5pt}
    \subfloat{
        \begin{minipage}[h]{0.19\textwidth}
    \includegraphics[width=0.95\textwidth,height=0.75\textwidth]{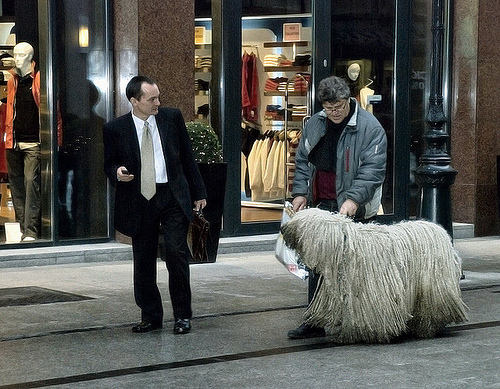}
    \includegraphics[width=0.95\textwidth,height=0.75\textwidth]{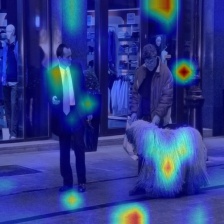}
    \includegraphics[width=0.95\textwidth,height=0.75\textwidth]{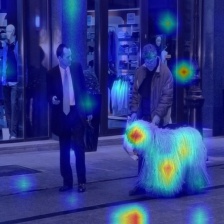}
    \includegraphics[width=0.95\textwidth,height=0.75\textwidth]{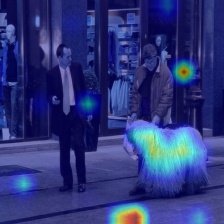}
    \includegraphics[width=0.95\textwidth,height=0.75\textwidth]{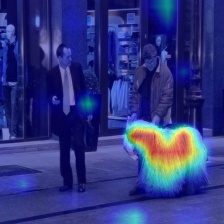}
    \centerline{Komondor}
    \label{fig:short-4d}
        \end{minipage}}
        \hspace{-5pt}
    \subfloat{
        \begin{minipage}[h]{0.19\textwidth}
    \includegraphics[width=0.95\textwidth,height=0.75\textwidth]{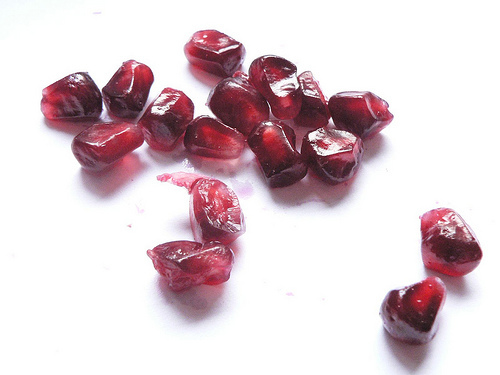}
    \includegraphics[width=0.95\textwidth,height=0.75\textwidth]{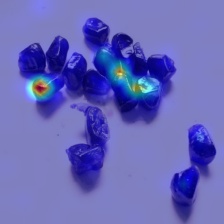}
    \includegraphics[width=0.95\textwidth,height=0.75\textwidth]{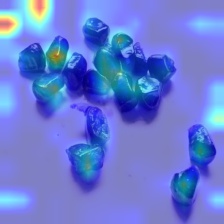}
    \includegraphics[width=0.95\textwidth,height=0.75\textwidth]{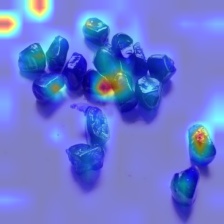}
    \includegraphics[width=0.95\textwidth,height=0.75\textwidth]{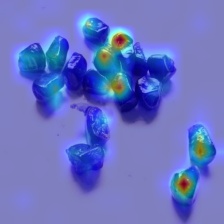}
    \centerline{Pomegranate}
    \label{fig:short-4e}
        \end{minipage}}
        \hspace{-5pt}
  \subfloat{
        \begin{minipage}[h]{0.19\textwidth}
    \includegraphics[width=0.95\textwidth,height=0.75\textwidth]{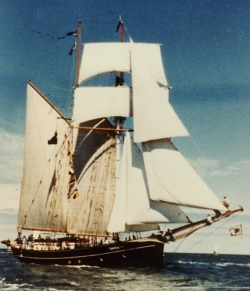}
    \includegraphics[width=0.95\textwidth,height=0.75\textwidth]{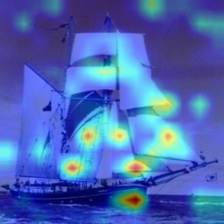}
    \includegraphics[width=0.95\textwidth,height=0.75\textwidth]{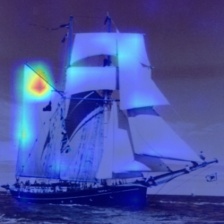}
    \includegraphics[width=0.95\textwidth,height=0.75\textwidth]{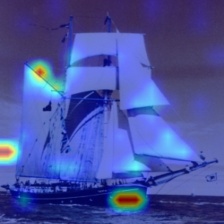}
    \includegraphics[width=0.95\textwidth,height=0.75\textwidth]{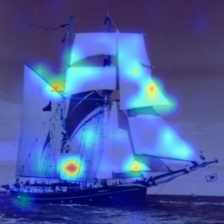}
    \centerline{Sailboat}
    \label{fig:short-4f}
        \end{minipage}}
      \end{center}
\caption{Visual explanations generated by a variety of pruned networks on the ILSVRC-12 validation set. From top to down: input image, visual explanation maps of the original DeiT-S, the pruned models by WDPruning \cite{0004HWCCC22}, UVC \cite{YuCSYTY0W22}, and our X-Pruner, respectively.}
\label{fig3}
\end{figure*}

\begin{figure}
  \setlength{\abovecaptionskip}{0cm}
  \setlength{\belowcaptionskip}{-0.2cm}
  \begin{center}
  \includegraphics[width=0.95\linewidth, height=0.4\linewidth]{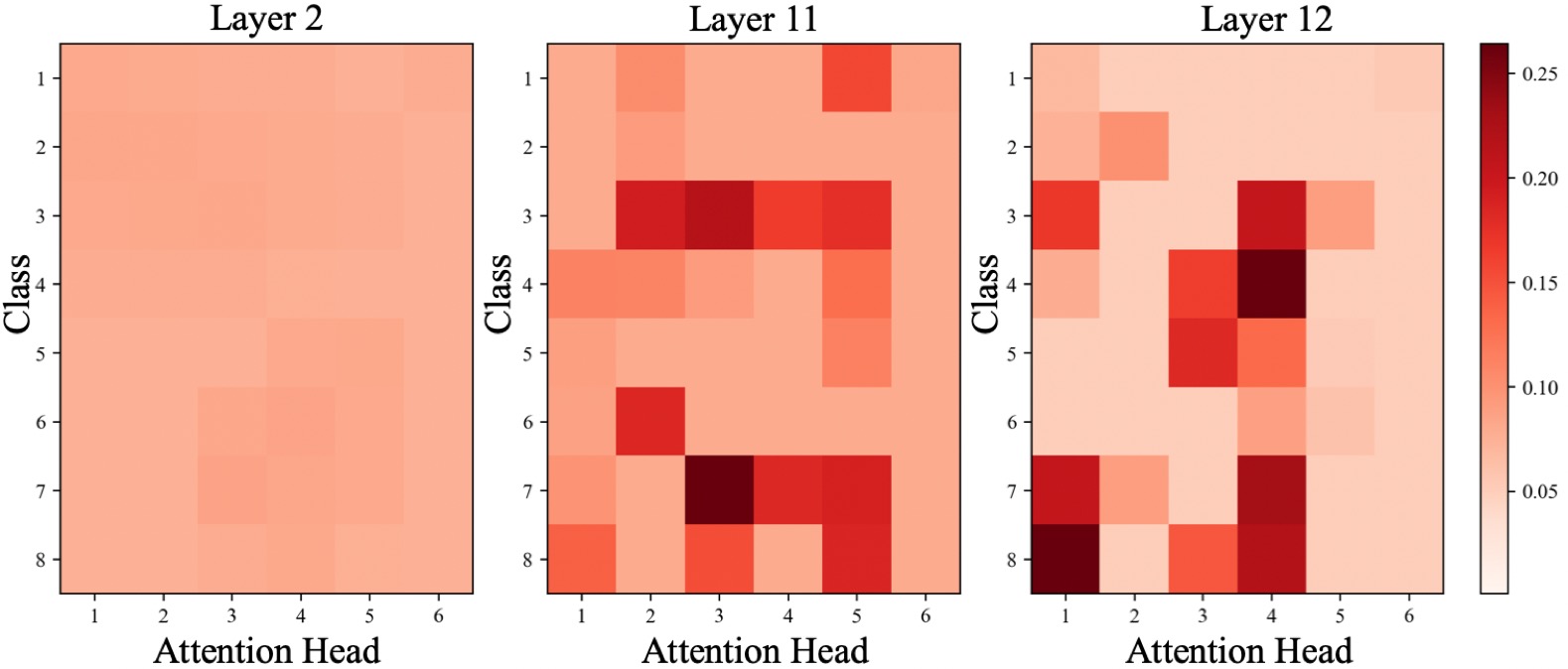}
  \end{center}
\caption{Explainability-aware mask values in varying layers for DeiT-S.}
\label{fig2}
\end{figure}

\subsection{Main results}
\cref{tab1} shows the superiority of X-Pruner over other state-of-the-art methods on ILSVRC-12. We observe that most existing pruning methods cannot provide noticeable FLOP savings without too much accuracy degradation. Instead, by learning the differentiable explainability-aware masks, our X-Pruner can reduce the computational costs by 51.3\%-66.1\% with much lower accuracy drops (0.72\%-1.1\%). Specifically, when pruning the DeiT-T, compared with WDPruning \cite{0004HWCCC22} that can only save 46.2\% FLOPs, it is observed that our proposed X-Pruner achieves much larger FLOP saving (66.1\% vs. 46.2\%) with less accuracy degradation (1.1\% vs. 1.86\%). For the larger model DeiT-S, while UVC \cite{YuCSYTY0W22} achieves the state-of-the-art top-1 accuracy among the existing methods, which is 78.82\% with a 49.6\% reduction in FLOPs, the X-Pruner reduces the FLOPs by 51.3\% while obtaining the top-1 accuracy of 79.04\%. These results demonstrate that the proposed X-Pruner outperforms existing pruning methods with more compact model sizes and better performance.

Meanwhile, we investigate the efficacy of our proposed method on another popular transformer, \ie, Swin Transformer \cite{liu2021swin}. The experimental results are presented in \cref{tab2}. For the Swin-T, the X-Pruner yields significantly better top-1 accuracy with substantially fewer FLOPs. More specifically, our method obtains 28.9\% FLOPs saving, and the top-1 accuracy only drops by 0.5\%. When pruning the Swin-S, compared to the state-of-the-art method WDPruning \cite{0004HWCCC22} which considers the dimensions for pruning, our X-Pruner also shows impressive superiority thanks to the use of the explainability-aware mask.

\begin{table}
  \setlength{\abovecaptionskip}{-8pt}
  \setlength{\belowcaptionskip}{0pt}
  \caption{Pruning results of Swin Transformer on the ILSVRC-12 dataset.} 
  \begin{center}
  \begin{tabular}{clcccc}
  \hline  
  & \makebox[0.05\linewidth][l]{Method} & \makebox[0.05\linewidth][c]{Top-1} & \makebox[0.08\linewidth][c]{FLOPs} & \makebox[0.14\linewidth][c]{Top-1 $\downarrow$} & \makebox[0.15\linewidth][c]{FLOPs $\downarrow$ (\%)} \\
  \hline
  \multirow{4}{*}{\rotatebox{90}{Swin-T}} & Baseline & 81.2 & 4.5 & 0.0 & 0.0 \\
  & STEP \cite{LiCS21} & 77.2 & 3.5 & 4.0 & 22.2 \\
  & ViT-Slim \cite{Chavan2022} & 80.7 & 3.4 & 0.5 & 24.4 \\
  & X-Pruner (Ours) & \bf{80.7} & \bf{3.2} & \bf{0.5} & \bf{28.9} \\
  \hline
  \multirow{4}{*}{\rotatebox{90}{Swin-S}} & Baseline & 83.2 & 8.7 & 0.0 & 0.0 \\
  & STEP \cite{LiCS21} & 79.6 & 6.3 & 3.6 & 27.6 \\
  & WDPruning \cite{0004HWCCC22} & 81.8 & 6.3 & 1.4 & 27.6 \\
  & X-Pruner (Ours) & \bf{82.0} & \bf{6.0} & \bf{1.2} & \bf{31.0} \\
  \hline
  \end{tabular}
  \end{center}
\label{tab2}
\end{table}

\subsection{Visualization and analysis}
We visualize the class-level visual explanation maps based on the DeiT-S as well as its pruned models by the LRP-based relevance method \cite{AbnarZ20}. \cref{fig3} provides a visual comparison based on randomly chosen ILSVRC-12 validation images. As can be seen from the figure, most of the visual explanation results of the full model still appear noise-like patterns to humans. However, the maps produced on the pruned model obtained by WDPruning \cite{0004HWCCC22} and UVC \cite{YuCSYTY0W22} are distorted. Though the predictions of the pruned models are correct, they produce incorrect explanation maps after the pruning process. Instead, we observe that the visual explanation maps produced on the pruned model of our X-Pruner are more compact and contain less noise.

Moreover, the learned mask values of attention layers shown in \cref{fig2} demonstrate that the proposed X-Pruner discovers the head importance appropriately without per-layer pruning ratio. Notably, the masks at higher layers (Layers 11 and 12) have higher values compared to the masks in Layer 2. Which indicates that in transformers, the lower layers attend to both local and global information, whereas the higher layers attend to global information. Thus rich semantic-level features are captured at higher layers, which are essential for the final predictions.

\begin{figure}
  \setlength{\abovecaptionskip}{0cm}
  \setlength{\belowcaptionskip}{-0.2cm}
  \begin{center}
  \includegraphics[width=0.98\linewidth, height=0.74\linewidth]{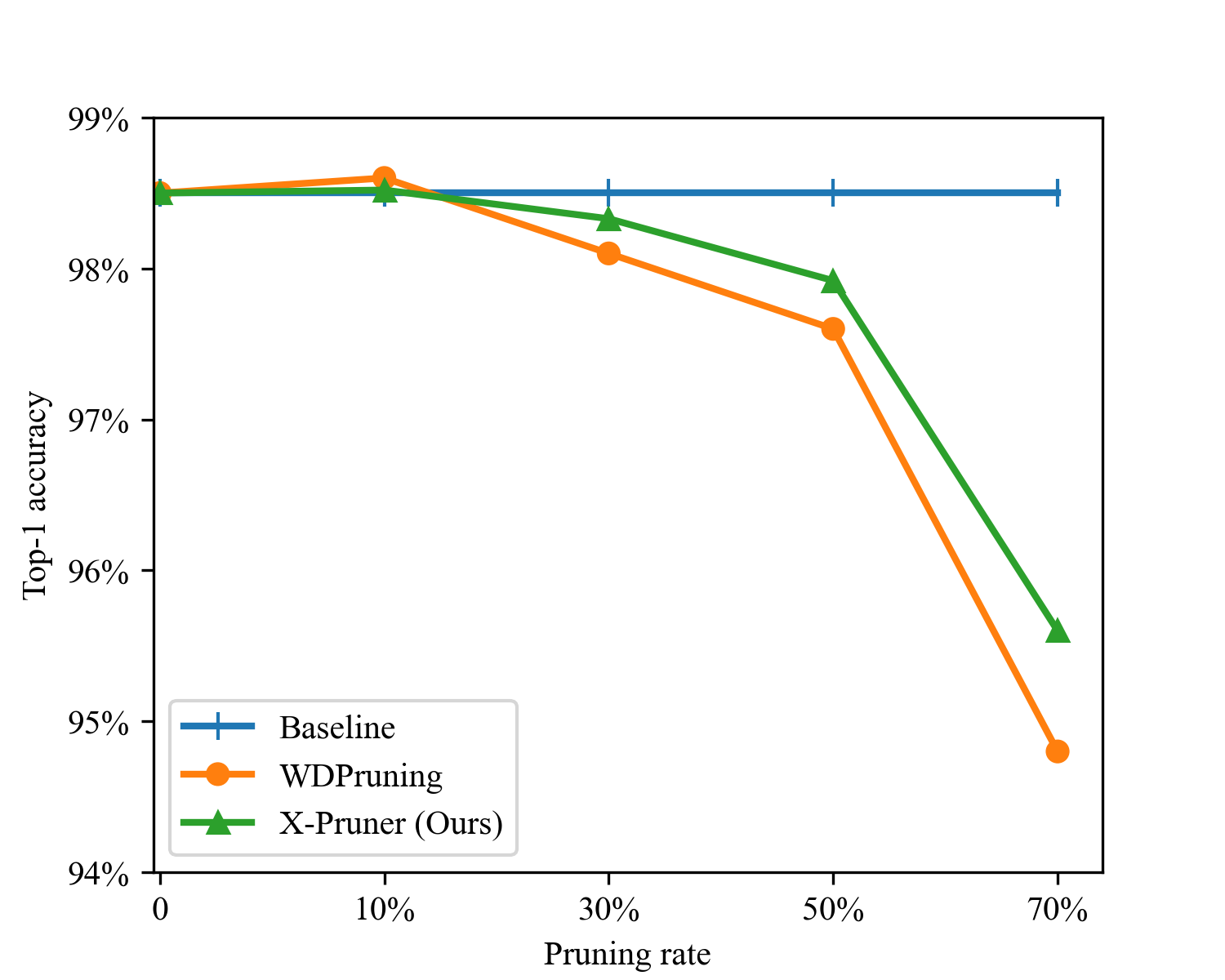}
  \end{center}
\caption{Top-1 accuracy for DeiT-S on CIFAR-10 with various pruning rates. "Baseline" denotes the unpruned baseline model.}
\label{fig4}
\end{figure}

We further compare our X-Pruner with the state-of-the-art method WDPruning \cite{0004HWCCC22} on CIFAR-10. \cref{fig4} depicts the top-1 accuracy of the DeiT-S with various pruning rates. As can be seen from the figure, at lower pruning rates, e.g., 10\%, both methods achieve slightly higher accuracy compared to the baseline. When it comes to larger pruning rates, compared to WDPruning \cite{0004HWCCC22}, our X-Pruner suffers less accuracy loss with the same pruning rates (e.g., 50\% or 70\%).

\begin{table}
  \setlength{\abovecaptionskip}{0cm}
  \setlength{\belowcaptionskip}{-0.1cm}
  \begin{center}
\begin{tabular}{lcc}
  \hline 
  \makebox[0.1\textwidth][l]{Setting} & \makebox[0.1\textwidth][c]{Top-1 $\downarrow$ (\%)} & \makebox[0.1\textwidth][c]{FLOPs $\downarrow$ (\%)} \\
  \hline
  w/o mask & 2.65 & 28.9 \\
  \hline
  w/o $\mathcal{L}_{\rm smooth}$ & 1.02 & 29.3 \\
  \hline
  w/o $\mathcal{L}_{\rm sparse}$ & 1.92 & 29.1 \\
  \hline
  X-Pruner & \bf{0.51} & \bf{28.9} \\
  \hline
  \end{tabular}
 \end{center}
\caption{Main results for pruning Swin-T under different configurations on ILSVRC-12.}
\label{table4}
\end{table}

\subsection{Ablation studies}
In this subsection, we first evaluate the effectiveness of explainability-aware masks in our proposed method based on the Swin-T model. \cref{table4} shows the detailed results, all of which are pruned using similar FLOPs pruning rates for a fair comparison. We first employ a class-agnostic strategy to train the explainability-aware mask, denoted as a w/o explainability-aware mask. That is, use the same mask for all the input given different classes. However, this strategy causes serious performance degradation (2.65\%) since it loses the class-wise signal to identify each unit's contribution. We further explore the impact of optimization constraints. Moreover, as is observed from \cref{table4}, when the masks are trained without the sparrse regularizer $\lambda_{\rm sparse}$, the trained model suffers a drop of 1.92\% in top-1 accuracy. Which proves our method effectively alleviates the problem of over-fitting and improves the performance. Finally, if the smooth constraint $\lambda_{\rm smooth}$ is removed, the top-1 accuracy is decreased by 1.02\%. Overall, our proposed proposed method X-Pruner is able to prune models effectively with desirable accuracy.  

\begin{figure}
    \setlength{\abovecaptionskip}{0cm}
    \setlength{\belowcaptionskip}{-0.2cm}
    \begin{center}
    \includegraphics[width=1\linewidth, height=0.76\linewidth]{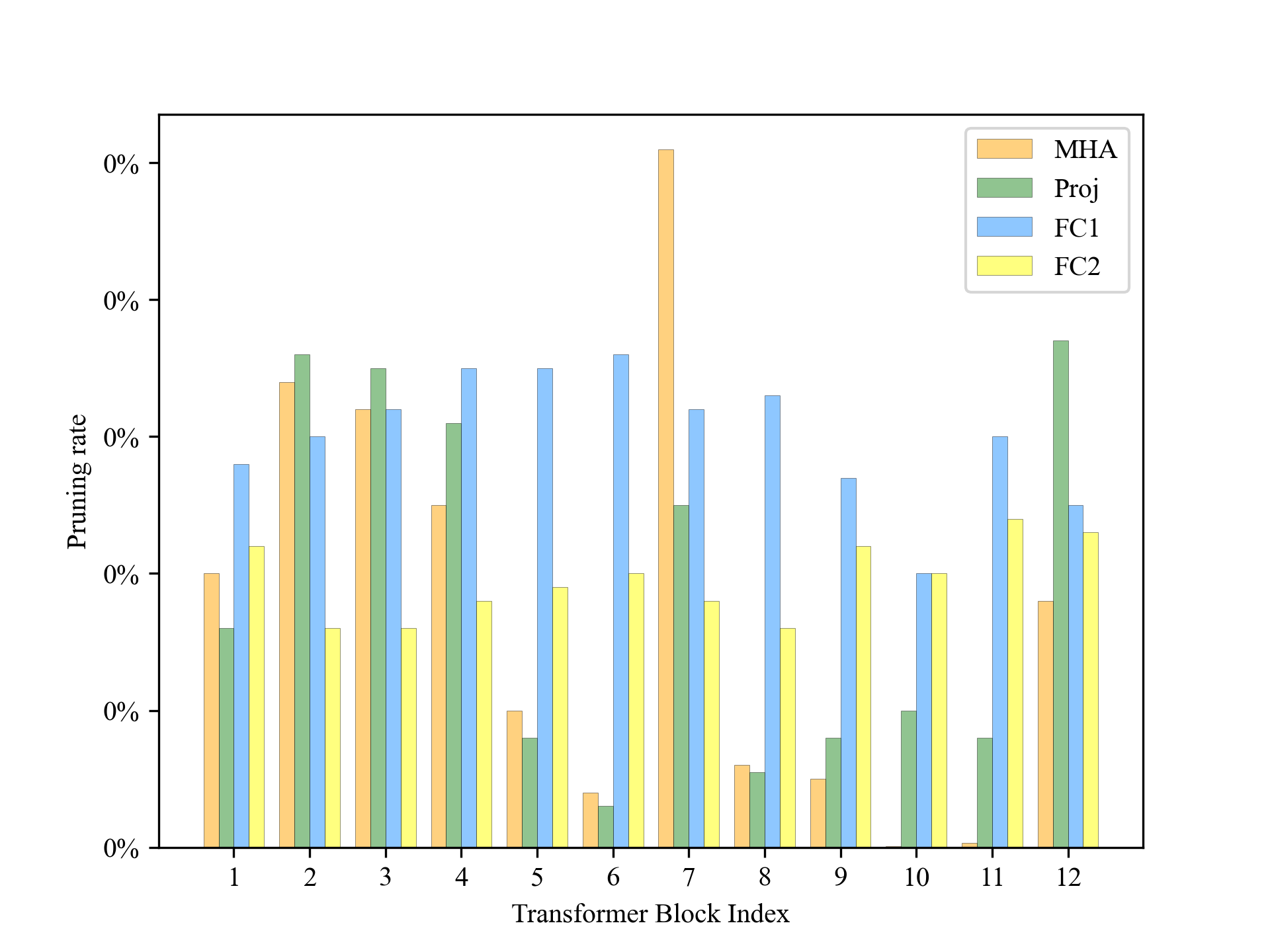}
    \end{center}
  \caption{The pruning rate of units on each block when the pruning rate is set at 0.3 for DeiT-S.}
\label{fig5}
\end{figure}

\begin{table}
    \setlength{\abovecaptionskip}{0cm}
    \setlength{\belowcaptionskip}{-0.2cm}
    \begin{center}
    \begin{tabular}{lcc}
    \hline  
    Method & Top-1 $\downarrow$ (\%) & FLOP $\downarrow$ (\%) \\
    \hline
    Random pruning & 2.28 & 47.2 \\
    \hline
    Uniform pruning & 4.05 & 47.4 \\
    \hline
    X-Pruner & \bf{0.87} & \bf{47.9} \\
    \hline
    \end{tabular}
\end{center}
    \caption{Main results of learnable pruning rate on DeiT-S.}
  \label{tab5}
  \end{table}

In \cref{tab5}, we further investigate the layer-wise pruning rate on ILSVRC-12 and compare it with both random pruning and uniform pruning. In our method, the number of pruned units for each individual layer is determined adaptively according to the global budget. The top-1 accuracy of uniform pruning is decreased by 4.05\%. We also apply the random pruning to the DeiT-S, which also achieves an inferior performance. Lastly, our proposed X-Pruner outperforms these two methods with minor top-1 accuracy drop (0.87\%). 

\begin{figure}
  \setlength{\abovecaptionskip}{0cm}
  \setlength{\belowcaptionskip}{-0.2cm}
  \begin{center}
  \includegraphics[width=1\linewidth, height=0.85\linewidth]{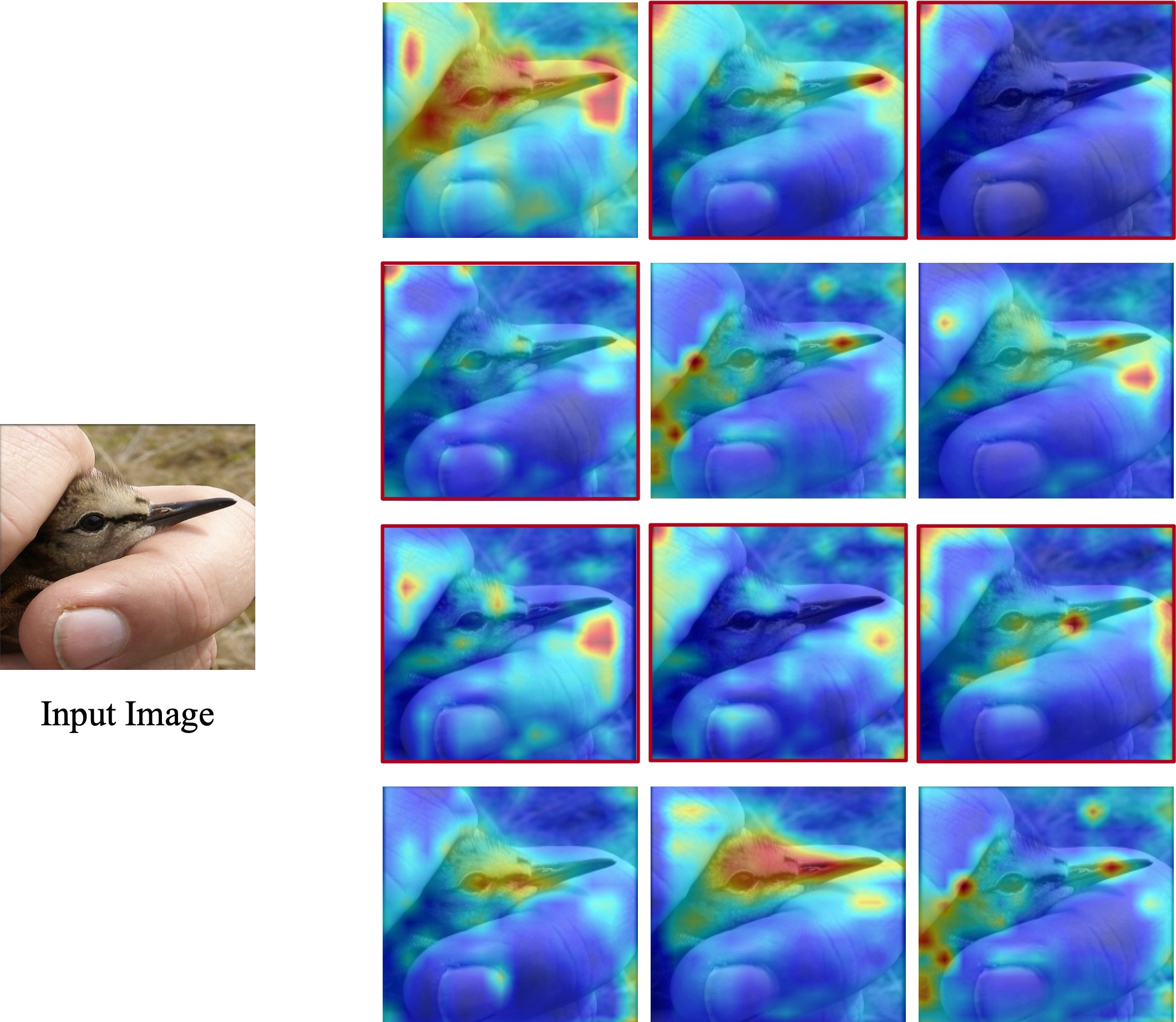}
  \end{center}
\caption{Visualization of the attention maps produced by the $4$-th layer for DeiT-B. Red box means the head is pruned based on our learned mask values.}
\label{fig6}
\end{figure}

We visualize the layer-wise pruning rate for the DeiT-S in \cref{fig5}. We observe that our method automatically learns the pruned architecture by taking into account the explainability-aware mask values, which is superior to estimating the importance of individual prunable units. Moreover, by visualizing the attention maps produced by the 4-th layer in DeiT-B model in \cref{fig6}, we observe that the proposed X-Pruner indeed removes the redundant heads that mainly focus on background and contribute less to the final prediction.

\section{Conclusion}
\label{sec5}
In this paper, we proposed the X-Pruner, a novel explainable transformer pruning framework. In X-Pruner, a novel explainability-aware mask is proposed to evaluate each prunable unit's contribution to predicting every class, which is fully differentiable and learned with a proposed class-wise regularizer to mitigate over-fitting. Then, a new explainable pruning process was introduced to learn layer-wise pruning rate until a resource constraint is reached. Extensive experiments demonstrate that the X-Pruner is able to significantly reduce the computational costs of several transformers in terms of model explainability. Moreover, it surpasses the state-of-the-art pruning methods with a minor accuracy drop.

{\small
\bibliographystyle{ieee_fullname}
\bibliography{mybib}
}

\end{document}